# Traffic Scene Small Target Detection Method Based on YOLOv8n-SPTS Model for Autonomous Driving

Songhan Wu[1]

[1]Department of Electrical Engineering and Computer Science, University of Michigan
wuumaa@umich.edu

*Abstract*—Against the backdrop of rapid technological development, autonomous driving technology is gradually moving from theory to practical application. The successful implementation of obstacle avoidance technology relies on accurate dynamic perception. This paper focuses on the key issue in autonomous driving: small target recognition in dynamic perception. Existing algorithms suffer from poor detection performance due to problems such as missing small target information, scale imbalance, and occlusion. This paper proposes an improved YOLOv8n-SPTS model, which significantly enhances the detection accuracy of small traffic targets. The innovations of the model mainly lie in three aspects: First, optimizing the feature extraction module. In the Backbone Bottleneck structure of YOLOv8n, 4 convolution modules are replaced with Space-to-Depth Convolution (SPD-Conv) modules. This module retains fine-grained information through space-to-depth conversion, reduces information loss in traditional convolution, and enhances the ability to capture features of low-resolution small targets. Second, enhancing feature fusion capability. The SPPFCSPC (Spatial Pyramid Pooling - Fast Cross Stage Partial Connection) module is introduced to replace the original SPPF module, integrating the advantages of multi-scale feature extraction from Spatial Pyramid Pooling (SPP) and the feature fusion mechanism of Cross Stage Partial Connection (CSP), thereby improving the model's contextual understanding of complex scenes and multi-scale feature expression ability. Third, designing a dedicated detection structure for small targets. A Triple-Stage Feature Pyramid (TSFP) structure is proposed, which adds a 160×160 small target detection head to the original detection heads to fully utilize high-resolution features in shallow layers; meanwhile, redundant large target detection heads are removed to balance detection accuracy and computational efficiency. Comparative experiments were conducted on the VisDrone2019-DET dataset, and the results show that YOLOv8n-SPTS ranks first in precision (61.9%), recall (48.3%), mAP@0.5 (52.6%), and mAP@0.5:0.95 (32.6%). Visualization results verify that the miss rate of small targets such as pedestrians and bicycles in occluded and dense scenes is significantly reduced. The model provides more accurate input for dynamic perception in complex environments of autonomous driving, improves the reliability of safe obstacle avoidance, and has important practical application value.

## I. Introduction

With the rapid development of social economy and the improvement of people's living standards, cars have become an indispensable part of modern families. Self-driving travel and freight transportation, as emerging modes of travel and logistics, are gradually becoming mainstream trends. This trend not only reflects people's pursuit of personalized and flexible travel modes but also their emphasis on time efficiency and economic benefits. However, with the surge in the number of vehicles, traffic safety issues have become increasingly prominent, with frequent traffic accidents posing a serious threat to people's lives and property safety. In addition, problems such as driver fatigue and inattention also seriously affect driving safety. Against this background, autonomous driving technology has emerged, which is regarded as an important way to solve current traffic problems due to its potential safety, efficiency, and comfort.

Autonomous driving technology enables vehicles to independently complete tasks such as environmental perception, decision-making planning, and vehicle control without direct intervention from human drivers by integrating advanced sensors, computer vision, artificial intelligence algorithms, and control systems. The realization of this technology can not only effectively reduce traffic accidents caused by human factors and improve road safety but also enhance traffic efficiency, reduce energy consumption, and mitigate environmental pollution by optimizing driving routes and behaviors. Meanwhile, autonomous driving can provide drivers with a more comfortable and convenient travel experience, especially in long-distance driving and freight transportation, which can significantly reduce the labor intensity of drivers and improve transportation efficiency.

Currently, in the practical application of autonomous driving deployment, there are remote areas or places where Vehicle to Everything (V2X) infrastructure is not yet popularized, and the safe obstacle avoidance of autonomous vehicles in obstacle blind-spot scenarios is not guaranteed [1]. These challenges are not only reflected in the difficulty of quantitatively evaluating potential collision risks caused by accidentally discontinuous information during driving and conducting safe and effective trajectory planning [2] but also closely related to the accuracy of small traffic target detection [3]. In complex traffic scenes, small target detection is of great significance to vehicle safe driving and dynamic perception. Due to the complexity of traffic scenes, occlusion between pedestrians and vehicles, and the small size of small targets in distant images, problems such as scale imbalance, occlusion in dense scenes, and blurred image resolution under different light conditions [4] affect the detection accuracy of small targets and exacerbate the safety risks of autonomous driving.

In recent years, deep learning-based target detection algorithms have made significant progress, but the research progress of small target detection has been relatively slow. Although substantial progress has been made in general target detection, the performance of small target detection is still far lower than that of medium and large targets. For example, on the COCO dataset, the mean Average Precision (mAP) of state-of-the-art detectors for small targets is only 28.3%, which is significantly lower than that for medium and large targets. This performance gap is mainly due to the difficulty in extracting information from small targets, their susceptibility to false detection, and the scarcity of large-scale datasets for small target detection. Based on existing research, this paper aims to better detect small traffic target to meet the safe obstacle avoidance needs of autonomous vehicles in complex traffic environments. By improving the small target detection algorithm, the model's ability to recognize occluded and small-sized targets is enhanced to provide more accurate input information for path planning.

## II. Related Work

### A. Target Detection Algorithms

Before the popularization of deep learning methods, target detection methods based on traditional image processing and machine learning dominated. These methods usually relied on manually designed feature extraction techniques combined with classifiers to achieve target detection. In 1999, a team led by Professor Geoffrey Hinton, one of the pioneers in the field of artificial intelligence, created a target detection algorithm based on Convolutional Neural Networks (CNN). This algorithm was the first deep learning method capable of completing vehicle target detection, laying the foundation for solving target detection problems in the field of computer vision [5]. Although the computing power of CNN was limited at that time, this initiative marked the initial exploration of the potential of deep learning methods in image analysis.

In 2017, Joseph Redmon and Ali Farhadi improved YOLOv1 and proposed YOLOv2, which adopted a deeper network architecture (Darknet-19) and new technologies such as Batch Normalization to further improve detection accuracy and speed [6]. YOLOv3 was also proposed by them in 2018. Based on YOLOv2, it used a deeper network structure (Darknet-53), enhanced the ability to detect small objects, and adopted a multi-scale detection strategy to improve performance in various detection tasks [7]. YOLOv4 introduced many new technologies, such as CSPDarknet53 as the backbone network, data augmentation, and YOLO-specific attention mechanisms, to improve detection accuracy and optimize computational efficiency. YOLOv5 was released by the Ultralytics team (including Glenn Jocher) in 2020. Although not recognized by Joseph Redmon, it quickly became one of the most popular versions in the YOLO series. This version focuses on model lightweight and ease of use, and provides multiple versions of different sizes (YOLOv5s, YOLOv5m, YOLOv5l, YOLOv5x) for different application scenarios (such as embedded devices) [8]. YOLOv6 was released by the Meituan team (including Xie Xue et al.) in 2022. While improving model accuracy, it specifically optimized detection speed and multi-scale target detection capabilities, making it suitable for real-time applications and low-power devices [9]. YOLOv7 further optimized the detection accuracy and training speed of the model, integrated new technologies such as transformer mechanisms, and improved performance in various practical applications, especially in small object detection and complex backgrounds. YOLOv8 was released by the Ultralytics team in 2023, emphasizing accuracy, speed, scalability, and multi-task capability. Compared with previous versions, YOLOv8 not only made significant progress in target detection accuracy but also enhanced adaptability to small objects, complex scenes (such as low light, occlusion, etc.), and multi-tasks (such as instance segmentation, key point detection, etc.) [10]. YOLOv8 specifically optimized the adaptive training mechanism, enhanced cross-domain adaptability and real-time detection capabilities, and further expanded the application scope of the model.

### B. Multi-Target Tracking Algorithms

Target detection mainly focuses on detecting objects in static images, but in the process of vehicle detection, dynamic targets need to be detected. Therefore, multi-target tracking technology has become a current research hotspot. In 2016, Alexandra Bewley et al. proposed an efficient and easy-to-implement multi-target tracking algorithm, Simple Online and Realtime Tracking (SORT) [11], which associates the Kalman filter with the Hungarian algorithm to complete target detection. The main advantages of the SORT algorithm are its high computational efficiency and strong real-time performance, making it suitable for low-latency real-time scenarios. It can handle rapidly changing environments and the detection of multiple targets, but in complex environments, especially when targets are occluded for a long time or their appearance changes, it is prone to target loss or incorrect matching. To address the limitations of SORT in complex scenes, Deep Learning-based SORT (DeepSORT) was proposed in 2017 [12]. DeepSORT made the following important improvements on the original SORT framework: On the one hand, in appearance feature extraction, the appearance features of each target are extracted through a convolutional neural network. Specifically, a deep learning module is introduced, using a pre-trained network (such as a ReID network) to extract the appearance features of each target, which are used for target re-identification and association when targets are occluded or interacting. On the other hand, improvements are made in data association. Traditional SORT only relies on target motion information for association, which is prone to association errors under complex conditions such as target overlap, occlusion, or interaction. DeepSORT combines appearance features (such as color, texture, etc.) with motion information (such as position, speed, etc.) and enhances the ability to maintain target identity by calculating the similarity between appearance features, solving problems such as occlusion and overlap [13]. Although deep learning-based multi-target tracking methods have made significant progress in accuracy and efficiency, they still face many challenges in practical applications. First, target occlusion and interaction remain difficult problems in multi-target tracking, especially in high-density target scenes, where tracking algorithms are prone to target loss or false matching. Second, real-time requirements are also an important issue faced by multi-target tracking, especially in real-time scenarios such as autonomous driving, which require more efficient algorithms to meet real-time needs. In addition, issues such as cross-scene adaptability and the diversity of training data also put forward higher requirements for the robustness of multi-target tracking.

## III. MODELING ALGORITHMS

### A. Overall Model Framework

In small traffic target detection, due to factors such as light and target scale, detection accuracy is low, and even missed detection and false detection may occur. Small targets cover a small area in the image, and the positioning error of their bounding boxes is higher than that of medium and large-scale targets. During training, the number of anchor boxes matching small targets is much lower than that of medium and large-scale targets, which further leads detection models to focus more on detecting medium and large-scale targets, making small target detection difficult. Moreover, small targets have low resolution and little visual information in images, making it difficult to extract discriminative features, and they are highly susceptible to environmental interference, so detection models cannot accurately locate and recognize small targets.

To improve the performance of YOLOv8n in small target recognition in complex traffic scenes, this paper makes targeted modifications to it, mainly including the following three aspects: (1) Small targets are small in size and difficult to detect. In the convolution layer of the Bottleneck structure of YOLOv8n, the second convolution module is replaced with a SPD-Conv module, while we also append SPD-Conv after each of the next three Conv layers. This module is specially designed to enhance the information detection ability in the feature extraction stage and reduce information loss, thereby improving the recognition accuracy of small targets. (2) Complex environment analysis is difficult. To further improve feature fusion efficiency and global context capture ability, this paper introduces the Cross Stage Partial (CSP) connection structure while maintaining the SPPF multi-scale pooling structure. This structure helps the model better understand and process contextual information in images, especially in complex traffic scenes. (3) Small targets have many types and are difficult to classify. Adding a small target detection layer: Finally, to enhance the model's ability to detect small targets, a dedicated small target detection layer is added to the head structure. This layer is specifically used to improve the expression ability of multi-scale features, enabling the model to pay more attention to and accurately recognize small-sized objects.

This paper constructs an optimized YOLOv8n-SPTS framework, as shown in Figure 1. This framework, while ensuring the efficient detection characteristics of the original YOLOv8n, is specially optimized for small target recognition. Specifically, the framework enhances the detection ability of small targets through the following methods: feature extraction optimization, where the framework can more effectively extract and retain feature information of small targets through the SPD-Conv module; feature fusion enhancement, where the framework achieves more efficient feature fusion using the CSP structure, improving the ability to understand complex scenes; multi-scale detection improvement, where the added small target detection layer makes the framework more excellent in multi-scale feature expression, especially enhancing the detection ability of small targets. The proposed YOLOv8n-SPTS model not only retains the efficient detection characteristics of YOLOv8n but also achieves significant improvement in the recognition performance of small targets in complex traffic scenes.

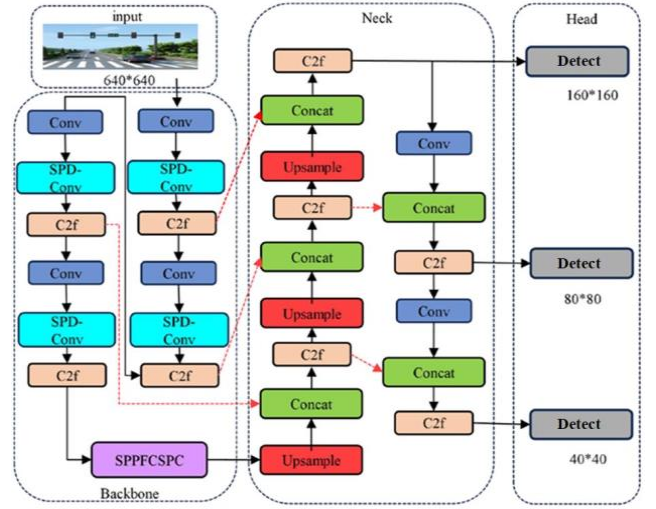

Fig. 1. YOLOv8n-SPTS framework model

### B. Model Improvement Details

SPD-Conv is an improved convolutional neural network module, consisting of two main parts: a Space-to-Depth (SPD) layer and a Non-Strided Convolution layer. It aims to solve the problem of information loss caused by strided convolution and pooling operations in traditional convolution operations, thereby improving the model's ability to detect low-resolution and small-sized targets. The SPD-Conv module adopted in this paper combines the advantages of the SPD layer and the non-strided convolution layer, reducing the spatial dimension of the feature map while retaining more spatial information, thus preserving richer fine-grained information.

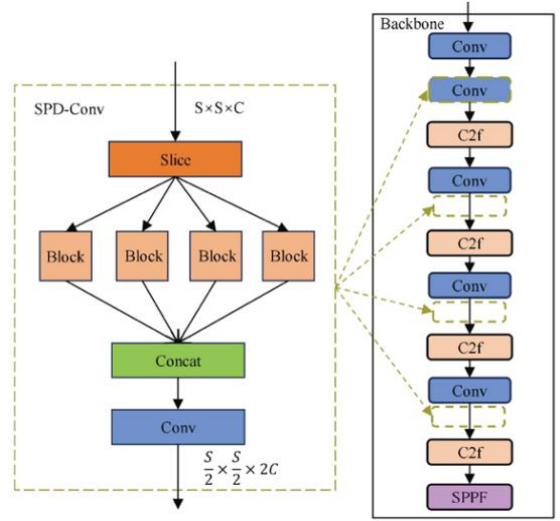

Fig. 2. Schematic diagram of improvements to the Backbone

This improvement not only significantly enhances the model's ability to recognize low-resolution and small-sized targets but also enables the model to effectively capture details of small objects even at small spatial resolutions, especially in traffic scenes, where it can better detect small targets such as pedestrians and bicycles. In the Backbone structure of the YOLOv8n model, except for the second Conv layer which is replaced with an SPD-Conv module, SPD-Conv modules are inserted at other selected positions. This improvement not only enhances the feature extraction ability of the model but also further improves the detection accuracy and robustness of the model for small targets through multi-level SPD-Conv modules, thus providing a more reliable solution for small target detection in traffic scenes. Through this improvement, the model can more accurately recognize and locate small targets in complex backgrounds, while reducing missed detection and false detection caused by information loss, providing important guarantees for the safety and reliability of autonomous driving systems, as shown in Figure 2.

Based on the Backbone structure improved by SPD-Conv, this paper introduces the SPPFCSPC (Spatial Pyramid Pooling - Fast Cross Stage Partial Connection) module as a new spatial pyramid pooling module to further improve the model's multi-scale feature extraction ability and feature fusion efficiency. The SPPFCSPC module combines the advantages of Spatial Pyramid Pooling (SPP) and Cross Stage Partial (CSP) connection structures. Its core idea is to enhance the model's adaptability to complex scenes through multi-scale pooling operations and cross-stage feature fusion, as shown in Figure 3. In specific implementation, the SPPFCSPC module first performs multi-scale pooling operations on the input feature map through the SPP structure to extract feature representations of different scales, and then inputs the pooled feature map into the CSP structure for cross-stage partial connection processing. Through channel segmentation and feature fusion, the expression ability of feature information is further enhanced. This design enables the SPPFCSPC module to effectively extract multi-scale features and achieve efficient feature fusion while keeping the scale of the feature map unchanged, thus providing stronger feature representation ability for the model's target detection tasks in complex scenes.

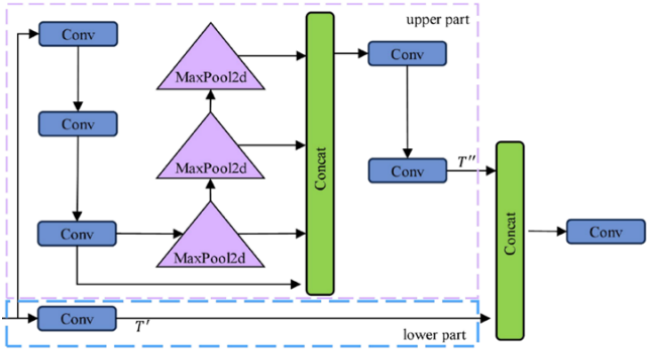

Fig. 3. Structural diagram of SPPFCSPC

The YOLOv8n model uses three detection heads with sizes of 20×20, 40×40, and 80×80 for target detection. However, in traffic scenes, the proportion of small targets is small, and the downsampling multiple of the original detection heads of YOLOv8n is large, leading to difficulty in deep feature maps effectively learning feature information of small targets, thus affecting the detection accuracy of small targets. To solve this problem, this paper proposes an enhanced feature pyramid structure, Triple-Stage Feature Pyramid (TSFP). This structure introduces a dedicated detection head for small target detection on the basis of retaining the original detection heads. By adding a detection head with 4× downsampling, it fully utilizes the rich detail information in shallow feature maps, thereby significantly improving the ability to capture features of small targets, as shown in Figure 4. The TSFP structure not only enhances the model's ability to detect small targets through multi-level feature fusion and optimized detection head design but also improves the computational efficiency of the model by removing redundant large target detection heads, providing an efficient and reliable solution for small target detection in traffic scenes.

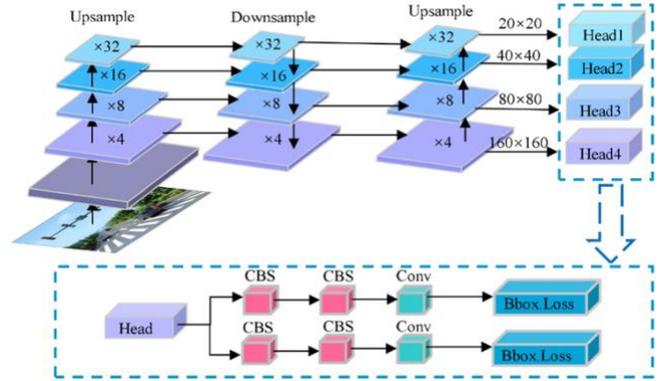

Fig. 4. Structural diagram of TSFP with additional small target detection head

IV. EXPERIMENTATION AND ANALYSIS

A. Experimental setup

The dataset used in this paper is the public small traffic target dataset VisDrone2019-DET. As shown in Figure 5, this dataset contains 10,209 high-resolution images, each with a resolution of 2000×1500, covering rich traffic scenes and diverse target categories. The dataset annotates 10 categories, including pedestrian, bicycle, people, car, tricycle, truck, van, bus, motor, and awning-tricycle. These categories cover common target types in traffic scenes, can effectively reflect the complexity and diversity of actual traffic environments, and provide comprehensive data support for model training and validation. The dataset is divided into a training set, a validation set, and a test set (test-dev). The training set contains 6,471 images, used for model parameter learning and feature extraction. Through large-scale data training, the model can learn feature representations and distribution rules of different targets; the validation set contains 548 images, used for adjusting model hyperparameters and evaluating model performance. The test set contains 1,610 images for final performance testing and algorithm comparison, and the results of the test set can objectively reflect the generalization ability and robustness of the model in practical applications. The high resolution and diverse target categories of the VisDrone2019-DET dataset make it an ideal choice for small traffic target detection research, effectively supporting the performance optimization and generalization ability improvement of the model in complex traffic scenes. In addition, complex traffic scenes (such as urban roads, intersections, highways) and different light conditions (such as daytime, nighttime, rainy days) included in the dataset further enhance the practicality and challenges of the dataset, providing important support for the performance optimization and generalization ability improvement of the model in real environments. The richness and diversity of small traffic objects from the VisDrone2019-DET dataset provide an important basis for the performance evaluation and optimization of the algorithm in this paper, and also lay data support for subsequent research and practical applications.

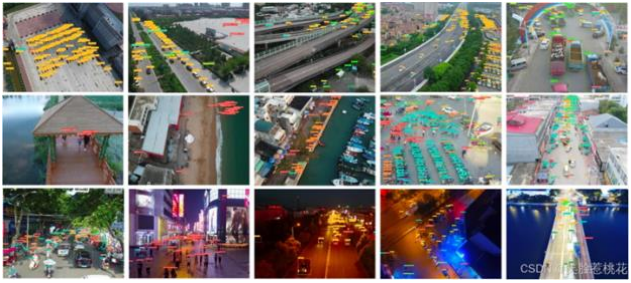

Fig. 5. VisDrone2019-DET dataset

The hardware environment used in this paper includes an Intel i7 10700 CPU, Windows 10 22H2 operating system, 45GB memory, and an NVIDIA RTX 3090 24GB GPU, providing strong computing support for model training and experiments. The software environment is based on Python 3.8, PyTorch 1.11.0, and CUDA 12.1 to ensure efficient operation of the algorithm and GPU acceleration. In comparative experiments, YOLOv3-tiny, YOLOv5, YOLOv6, YOLOv10, and YOLOv8n improvement experiments were all conducted in the same environment to ensure the fairness and comparability of experimental results. In terms of experimental parameter settings, the resolution of input images is 640×640, the number of training epochs is 300, the early stopping mechanism is set to 50 epochs, the optimizer adopts Stochastic Gradient Descent (SGD), and other parameters remain default.

### B. Experimental results

This paper uses P (Precision), R (Recall), mAP@0.5 (Mean Average Precision at IoU=0.5), and mAP@0.5:0.95 (Mean Average Precision at IoU=0.5:0.95) as indicators to evaluate model performance.

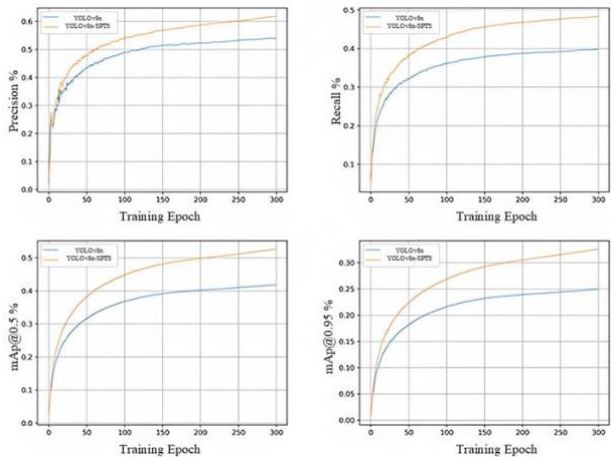

Fig. 6. Training curve

The training loss comparison of the model before and after improvement is shown in Figure 6. This paper compares the training accuracy of the YOLOv8n-SPTS model with and without SPD-Conv, SPPFCSPC and TSFP. It can be seen from the figure that the values of P, R, mAP@0.5, and mAP@0.5:0.95 of YOLOv8n-SPTS are all higher than those of the YOLOv8n model, indicating that the proposed model is superior to the original model in small target detection on the traffic image dataset.

The improved model YOLOv8n-SPTS in this paper is compared with mainstream target detection models such as YOLOv3-tiny, YOLOv5, YOLOv6, and YOLOv10, as shown in Table 1. YOLOv6 has the lowest precision, mAP@0.5 and mAP@0.5:0.95, indicating that the proportion of incorrect prediction samples of the model is high, and the model's detection performance is unstable and incomplete; YOLOv3-tiny has the lowest recall rate, proving that the model has a high miss rate and poor small target detection accuracy; The proposed YOLOv8n-SPTS model ranks first in all detection accuracy indicators. For example, in mAP@0.5, it has an increase of 10.8 percentage points compared with the original YOLOv8n model, and 11.2 percentage points higher than the newer YOLOv10n model. The experimental results effectively prove the effectiveness and application potential of the proposed model.

TABLE I. EXPERIMENTAL RESULTS

| Methods | P | R | mAP@0.5 | mAP@0.5:0.95 |
|---|---|---|---|---|
| YOLOv3-tiny | 58.7 | 32.9 | 36.2 | 21.5 |
| YOLOv5 | 51.7 | 38.2 | 39.6 | 23.6 |
| YOLOv6 | 48.3 | 34.6 | 35.2 | 21.0 |
| YOLOv8n | 54.1 | 39.8 | 41.8 | 24.9 |
| YOLOv10n | 53.7 | 39.6 | 41.4 | 24.8 |
| YOLOv8n-SPTS | **61.9** | **48.3** | **52.6** | **32.6** |

To more intuitively compare the performance gap between the model before and after improvement, this paper selects some test images for visualization comparison, and the results are shown in Figure 7.

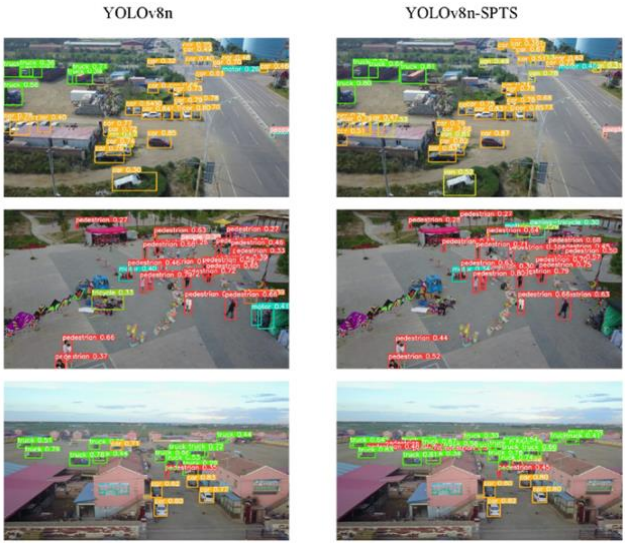

Fig. 7.  Visualization comparison

The improved YOLOv8n-SPTS model shows higher accuracy and lower miss rate in detecting small targets such as pedestrians and bicycles, especially in complex backgrounds and dense small target scenes, where the model can more accurately recognize and locate targets. Through visualization comparison, this paper verifies the effectiveness of the SPD-Conv, SPPFCSPC, and TSFP modules in improving small target detection performance, providing an efficient and reliable solution for small target detection in traffic scenes.

## V. Conclusion

To improve the current situation of missed detection and insufficient accuracy of traditional target detection algorithms in detecting small-sized targets and pedestrians, this paper chooses to optimize the fast detection model YOLOv8n and improve the model's detection performance in traffic scenes through multi-level improvements. First, the SPD-Conv module is integrated into the original convolution module. Using its space-to-depth conversion characteristic, information loss in traditional convolution operations is reduced, thereby enhancing the model's ability to capture detailed features of small targets, especially in detection tasks of low-resolution and small-sized targets. Second, the original SPPF module is replaced with the SPPFCSPC module, which combines the advantages of spatial pyramid pooling and cross-stage partial connection to further improve multi-scale feature extraction ability, enabling the model to more effectively process targets of different scales, especially feature information of small targets. These two improvements are implemented in the Backbone structure, significantly enhancing the feature extraction ability and robustness of the model. Finally, to address the problem of the small proportion of small targets in traffic images, this paper proposes a small target detection head TSFP structure, which is designed based on a three-layer feature pyramid. It not only enhances the feature fusion ability of FPN but also fully utilizes high-resolution information in shallow feature maps through the newly added 4× downsampling detection head, thereby significantly improving the model's sensitivity and detection accuracy to small targets. The TSFP structure effectively solves the missed detection problem in small target detection through multi-level feature fusion and optimized detection head design, and further reduces the model complexity through pruning operation to remove detection heads with redundant processing of large targets, achieving a balance between detection performance and computational efficiency. These improvements enable the model to more accurately recognize and locate small targets in traffic scenes, providing important guarantees for the safety and reliability of autonomous driving systems.